**Does *Machine* "know" interpersonal pragmatics? Evidence from MARBERT's learning of emoji pragmatics in Arabic digital discourse**

Mohammed Q. Shormani, Ibb University, Ibb, Yemen
shormani@ibbuniv.edu.ye/https://orcid.org/0000-0002-0138-4793

## Abstract

This study examines Transformer-based models' ability to learn emoji pragmatics in Arabic digital discourse (ADD) providing evidence from MARBERT's behavior with interpersonal pragmatic functions (IPFs). A corpus of 8504 unique emoji-posts collected from Facebook via Python was used in the study. These posts were manually annotated, developed and labeled for 5 IPFs: *Politeness, Respect, Solidarity, Empathy,* and *Encouragement*. A mixed-method approach was employed comprising statistical methods, and interpretative ones involving speech act theory, politeness theory, and rapport management theory. MARBERT was finetuned to model these context-dependent pragmatic functions. Findings unveil MARBERT's ability to learn such IPFs achieving strong performance on the unseen data: accuracy of 93%, a micro F1-score of 0.61, and a macro F1-score of 0.56, demonstrating its effectiveness in capturing interpersonal functions beyond conventional sentiment analysis. Function-level evaluation showed that Politeness and Respect were identified more accurately than Solidarity, reflecting differences in the explicitness and contextual dependence of IPFs. The study concludes that Transformer-based models learn patterns of face management and relational communication but face challenges by highly implicit social meanings. It contributes a novel computational approach for modeling emoji pragmatics and advances the integration of interpersonal pragmatics with NLP for digital communication research.

## 1. Introduction

It goes without saying that currently digital sphere has transformed online interaction into a predominantly multimodal phenomenon, where meaning is constructed through the integration of linguistic, visual, and technological resources (see e.g., O'Halloran et al., 2010; Satar et al., 2023; Moschini & Sindoni, 2021; Yus, 2011, 2014, 2025). These include the fact that emojis have become indispensable components of digital communication across social media platforms such as Facebook, X, Instagram. There arise several questions, given the development artificial intelligence models has undergone (see e.g., Shormani, 2025), concerning the possibility of predicting these digital pragmatics instances by Transformer-based models like BERT (see e.g., Devlin et al., 2019), MARBERT (see e.g., Abdul-Mageed & Elmadany, 2021), and XLM-Roberta (see e.g., Conneau et al.,2020; Salve & Tubil, 2025). This is perhaps due to the fact that computational research has largely remained confined to modeling emojis from semantic and affective perspectives. Recent advances in machine/deep learning and Transformer-based language models have substantially improved the automatic recognition of sentiment and emotion emojis across several languages. However, these approaches typically assume that emojis primarily encode emotional or lexical meaning, overlooking their crucial role in accomplishing interpersonal pragmatic functions (IPFs) during digital interaction (see e.g., Kralj Novak et al., 2015; Ahamad & Mishra, 2025; Zhao et al., 2018), and here the study problem lies.

Digital pragmatics, an emerging interdisciplinary field that investigates how pragmatic meaning is constructed, interpreted, and negotiated within digital interaction, corpus pragmatics, and sociocultural environments (Vásquez, 2022). Traditionally, pragmatics was understood as the examination of how language functions within specific contexts, looking specifically at how people convey meanings that stretch beyond the literal definitions of their words. It explores how situational context, a speaker's underlying goals, and social conventions help listeners deduce the intended message behind an utterance (see e.g., Yule, 2014). Within modern social media platforms, this field has evolved into digital pragmatics to address novel online communication challenges. This concept largely stems from *cyberpragmatics* introduced by Yus (2011) to study internet-mediated interactions. Since then, the term has been adopted for analyzing social media discourse (see e.g., Bou-Franch & Blitvich, 2018; Bou-Franch, 2020). Additionally, emojis have evolved into sophisticated pragmatic resources that shape interpersonal interaction, regulate social relationships, express attitudes, and facilitate contextual interpretation (see e.g., Yus, 2014, 2025; Caspi & Raz, 2025). Thus, recent developments in digital pragmatics increasingly conceptualize emojis not as decorative symbols or simple emotion indicators but as context-dependent semiotic resources whose communicative functions emerge through interaction, discourse context, and shared sociocultural knowledge (see e.g., Ge & Herring, 2018). From a digital-pragmatic perspective, emojis frequently perform relational work by expressing politeness, signaling respect, constructing solidarity, communicating empathy, and providing encouragement (Yus, 2025).

Although Arabic NLP has witnessed remarkable progress through the development of specialized Transformer models such as BERT, AraBERT, and MARBERT, research investigating emojis as markers of interpersonal pragmatics remains scarce (see e.g., Devlin et al., 2019; Abdul-Mageed & Elmadany, 2021). Existing studies overwhelmingly focus on sentiment analysis, emotion detection, or sarcasm identification, leaving the computational modeling of interpersonal meaning largely untouched (see e.g., Kralj Novak et al., 2015; Ahamad & Mishra, 2025). However, few studies have attempted to operationalize established pragmatic theories within deep learning frameworks for ADD (see e.g., Shormani et al., 2026). Addressing this gap requires moving beyond affective classification towards modeling the relational functions that emojis perform during naturally occurring online communication. Beyond sentiment analysis, etc., pragmatic interpersonal functions are inherently context-dependent and cannot be adequately explained through sentiment polarity or discrete emotion categories alone (cf. Albluwi, 2026).

With this in mind, this study investigates the IPFs of emojis in Arabic digital discourse (ADD) by finetuning the state-of-the-art MARBERT for multi-label classification. Drawing upon Transform-based studies (see e.g., Devlin et al., 2019; Abdul-Mageed & Elmadany, 2021), digital pragmatics (see e.g., Yus, 2011; Bou-Franch & Blitvich, 2018; Bou-Franch, 2020), speech act theory (see e.g., Austin, 1962), politeness theory (see e.g., Brown & Levinson, 1987), and rapport management theory (see e.g., Spencer-Oatey, 2000), the study operationalizes 5 IPFs, namely *politeness, respect, solidarity, empathy,* and *encouragement*, as theoretically motivated categories for computational modeling, specifically with MARBERT. By demonstrating that Transformer-based language models can learn these context-dependent interpersonal functions, this study extends current emoji research beyond conventional sentiment analysis towards the modeling of interpersonal meaning.

Thus, this article is organized as follows. Section 2 reviews the theoretical framework and literature. Section 3 outlines the methodological rigor, detailing the data collection procedures, platform selection criteria, and the discourse analysis approach employed. Section 4 presents the analysis and central findings, examining how users construct meaning, manage interpersonal relationships, and utilize contextual cues in online interactions. Section 5 discussers these results. Section 6 concludes the article summarizing the main contributions, addressing the study potential limitations, and suggesting implications for future research.

## 2. Theoretical foundations and literature review

### 2.1. Digital pragmatics

Digital pragmatics is considered a newly emerging concept in the study of digital communication. It has substantially transformed the "traditional view of pragmatics" into how language is produced, interpreted, and negotiated across digital platforms. The main concept of digital pragmatics is how to interpret compensation for face-to-face interaction, by involving the dynamic integration of linguistic, visual, and technological resources to accomplish communicative goals (see e.g., Vásquez, 2022; Bou-Franch & Blitvich, 2018; Bou-Franch, 2020). Digital platforms such as Facebook, X, Instagram, TikTok, have thus become multimodal interactional environments in which written language is routinely accompanied by emojis, stickers, GIFs, images, hyperlinks, hashtags, and other semiotic resources. and other platform-specific affordances (Hurley, 2019). Within these environments, communication extends beyond the transmission of propositional content to encompass the continuous negotiation of interpersonal relationships, identities, attitudes, and social actions through multiple interacting semiotic modes (see also Gibson et al., 2018).

These developments have given rise to digital pragmatics, most central to which is recognizing that communication has become inherently multimodal (O'Halloran et al., 2010; Satar et al., 2023). Current online discourse depends heavily on users combining diverse semiotic resources to achieve communicative objectives that would otherwise require prosody, gesture, gaze, or facial expression in face-to-face conversation. In this sense, emojis could be seen as multimodal resources, emerging as one of the most pervasive and communicatively significant components of online interaction. Their contribution extends far beyond emotional expression, a folded-hands emoji 🙏, for example, may express gratitude, respect, apology, supplication, or politeness depending entirely on the communicative context in which it occurs. Another example is the red heart emoji ❤️; it may index affection, solidarity, appreciation, encouragement, or agreement according to the ongoing interaction (see also Yus, 2025). Such variability illustrates that emoji interpretation is substantially pragmatic more than semantic, requiring inferential processes that integrate linguistic, contextual, interpersonal, and cultural information. The interpersonal dimension of emoji use is particularly significant (cf. Morsi, 2026). Digital interaction is inherently relational to establish rapport, negotiate identities, manage face, express alignment, and maintain social cohesion (Herring & Ge-Stadnyk, 2024; Yus, 2025). Emojis facilitate these relational processes by providing efficient visual mechanisms through which users communicate empathy, appreciation, encouragement, respect, affiliation, or politeness without interrupting the flow of textual interaction. This interpersonal orientation aligns closely with recent developments in digital discourse research, which increasingly conceptualize emojis as pragmatic markers of relational work rather than emotional icons.

In this study, we adopt this digital pragmatic perspective to conceptualize emojis as multimodal interpersonal resources whose primary communicative value lies in the negotiation of social relationships within ADD. Thus, we depart from considering emojis as indicators of sentiment or isolated emotional states, viewing them as context-dependent pragmatic devices that perform identifiable interpersonal functions during naturally occurring digital interaction. Specifically, it investigates 5 theoretically motivated interpersonal functions: *politeness, respect, solidarity, empathy,* and *encouragement* which represent distinct forms of relational work accomplished through emoji use. This perspective shifts computational emoji research beyond conventional sentiment analysis towards the modeling of interpersonal meaning, thereby establishing the theoretical foundation on which the subsequent discussion of speech acts, politeness, rapport management, and computational operationalization is developed.

## 2.2. Speech act theory

Perhaps the most important conception in Speech Act Theory (SAT) is "words/verbs as actions". It constitutes one of the foundational frameworks in pragmatics for explaining how language performs actions rather than merely conveying information (see e.g., Austin, 1962; Leech, 2014). For instance, Austin (1962) argues that speaking itself constitutes a form of social action through which individuals accomplish communicative purposes such as requesting, apologizing, congratulating, thanking, promising, warning, or expressing sympathy. In this sense, then, utterances do not primarily describe states of affairs, but rather language is inherently performative, and successful communication depends not only on the literal meaning of words but also on the interlocutor's communicative intentions and the hearer's pragmatic interpretation. It follows that SAT shifted the focus of pragmatic inquiry from sentence meaning to language use, emphasizing how speakers/writers accomplish interpersonal goals through contextually situated interaction (Yule, 2014).

According to Austin (1962, pp. 94-103, 150), there are 3 interrelated dimensions of every utterance: the locutionary act, referring to the literal linguistic expression; the illocutionary act, representing the communicative intention performed by the speaker; and the perlocutionary act, describing the effect that the utterance produces on the hearer. The illocutionary act occupies a central position because it reflects what speakers actually intend to accomplish through communication. For example, the statement *"Congratulations on having a new born!"* does more than communicate information; it performs the social act of congratulating an interlocutor on having a new baby. However, sometimes utterances like *"Sorry"* may communicate several recognizable communicative actions, depending substantially on pragmatic interpretation rather than semantic content alone. Put simply, the utterance *"Sorry"* could mean "please repeat what you have said!" when a hearer did not understand what a speaker said, "it's my mistake!", when a hearer apologized for making a mistake, "please move a little" when a speaker asks his/her interlocuter "to move away to pass", etc.

Given these communicative functions, and following Austin's seminal thoughts, Searle (1969, p. 10-13) developed a systematic classification of speech acts based on their communicative functions. He distinguished several broad categories, including representatives, "to commit the speaker (in varying degrees) to something's being the case, to the truth of the expressed proposition", ii) directives, whose "illocutionary point … consists (being) attempts … by the speaker to get the hearer to do something", iii) commissives, through which speakers commit themselves to future actions, iv) expressives, which communicate psychological states or social attitudes, and v) declarations, which bring about institutional or social changes through their very utterance. These communicative functions were actually developed to explain spoken language, but subsequently these speech acts are equally applicable to written communication, particularly in digitally environments where communicative intentions must often be inferred from contextual rather than prosodic cues (cf. Hancock et al., 2024). Given our example above, the utterance *"Sorry"* could mean "please move a little" when a speaker asks his/her interlocuter "to move away to pass", etc. In pragmatics, this sense indicates *politeness* mitigating a request, thus performing a directive speech act.

The emergence of digital communication has considerably expanded the scope of SAT by introducing multimodal resources that participate directly in the realization of communicative actions. Current digital interaction rarely relies exclusively on linguistic expressions. Instead, users combine written language with emojis, stickers, images, reactions, GIFs, and other visual resources that jointly contribute to the interpretation of communicative intentions (Sasamoto, 2023). Within digital pragmatics, speech acts thus emerge from the interaction between verbal and non-verbal semiotic resources rather than from textual language alone (see also Danesi, 2017). This makes it clear that emojis, one way or another, are related speech acts, as they "perform" or reinforce actions (Herring & Ge-Stadnyk, 2024). As reinforcing ingredients, the folded-hands emoji 🙏, for instance, may reinforce the illocutionary force of praying, thanking, requesting, apologizing, or expressing respect depending on the accompanying discourse. Similarly, a flexed-biceps emoji 💪 may strengthen acts of encouragement or motivation, while a crying-face emoji 😢 contributes directly to expressions of sympathy or condolence. As for performing actions, emojis function as independent speech acts without accompanying verbal language. A single thumbs-up emoji 👍, for instance, may perform the act of agreement or approval, agreement, encouragement (see also Ge & Herring, 2018; Yus, 2025).

In this study, we take emojis as pragmatic instances, performing speech acts and reinforcing illocutionary force. The 5 interpersonal pragmatics *politeness, respect, solidarity, empathy,* and *encouragement* are viewed as recurrent interpersonal speech acts realized through emoji-mediated interaction in ADD. This study thus interprets emojis as multimodal devices that perform, reinforce, or modify communicative actions directed towards maintaining interpersonal relationships in ADD.

**2.3. Politeness theory**

Politeness Theory (PT) was first developed by Brown and Levinson (1987) based on the basis of Goffman's (1955) notion of face. It proposes that communication extends beyond the exchange of information to encompass the continuous negotiation of individuals' public self-image during social interaction. PT constitutes one of the most influential frameworks in interpersonal pragmatics for explaining how speakers manage social relationships through language. In this sense, every communicative act involves the potential to either maintain or threaten interpersonal relationships, making politeness an essential mechanism for achieving successful communication. the concept of face refers to the socially recognized public self-image that individuals seek to maintain during interaction (cf. Morsi, 2026). PT distinguishes between 2 complementary dimensions of face: positive face represents an individual's desire to be appreciated, approved of, accepted, and recognized as a valued member of a social group. It reflects the fundamental human need for affiliation, solidarity, and interpersonal acceptance. Negative face, however, refers to an individual's desire for autonomy, freedom of action, and freedom from unnecessary imposition.

Successful interaction therefore requires speakers to balance these 2 dimensions while pursuing their communicative objectives (see e.g., O'Driscoll, 2011).

Given that communication frequently challenges these interpersonal needs, Brown and Levinson (1987) introduce the notion of face threatening acts, in which requests, disagreements, criticism, refusals, advice, corrections, or even expressions of disagreement may threaten either the positive or negative face of interlocutors. these strategies include positive politeness and negative politeness as particularly relevant for understanding interpersonal communication in digital environments (see also Conlan, 2005). Positive politeness seeks to reduce interpersonal distance by emphasizing shared identity, approval, cooperation, common ground, and group membership. Typical realizations include compliments, expressions of agreement, shared humor, encouragement, and demonstrations of empathy or emotional support (Haugh, 2013). However, negative politeness aims to respect the interlocutor's autonomy by mitigating imposition, expressing deference, apologizing, hedging requests, and acknowledging social distance. These strategies enable speakers to accomplish communicative goals while simultaneously maintaining interpersonal relationships. Brown and Levinson distinguish 4 broad strategic options, viz., i) performing the act baldly without mitigation, ii) employing positive politeness strategies, iii) employing negative politeness strategies, or iv) avoiding the act altogether through indirectness or silence. The specific strategy selected depends on contextual variables including social distance, relative power, and the perceived seriousness of the imposition.

In positive politeness, this study considers emojis to be contributing ingredients to the construction of interpersonal closeness by expressing politeness, encouragement, solidarity, empathy for interpersonal support. Emojis such as ❤️, 🤝, 🤗, or 🎉 reinforce shared identity and mutual affiliation, signaling that the speaker recognizes and values the interlocutor as a member of the same social community. However, these interpersonal functions, specifically politeness cannot be understood solely as an individual strategy for mitigating face threats but must instead be interpreted within broader social, cultural, and relational contexts (cf. Locher and Watts, 2005; Culpeper, 2011).

### 2.1.4. Rapport management theory

The first scholar to develop rapport management theory (RMT) is Spencer-Oatey (2000) perhaps to overcome the PT limitations as spotted above. Within this theory, she extends and refines classical politeness theory by proposing that interpersonal communication involves far more than the protection of face. While Brown and Levinson primarily explain how speakers mitigate face threatening acts during individual interactions, Spencer-Oatey argues that communicative behavior should be understood within the broader process of establishing, maintaining, strengthening, and repairing interpersonal relationships. RMT thus shifts the analytical focus from isolated politeness strategies towards the continuous negotiation of social relationships, interpersonal expectations, and communicative harmony. Its central contribution to RMT is its multidimensional conceptualization of interpersonal relations. Spencer-Oatey argues that rapport is jointly determined by 3 interconnected components: i) face, ii) sociality rights and obligations, and iii) interactional goals. As for face, Spencer-Oatey differentiates between quality face and identity face. Quality face concerns an individual's desire to have their personal qualities, competence, abilities, and achievements positively evaluated by others. Identity face, however, relates to the recognition of socially significant roles, group memberships, institutional positions, and public identities (see also Spencer-Oatey, 2008; Spencer-Oatey & Wang, 2026).

These 2 dimensions are particularly relevant in digital communication, where users frequently acknowledge expertise, celebrate achievements, recognize authority, or reinforce professional identity through multimodal interaction. Regarding sociality rights and obligations component, it refers to individuals' expectations regarding appropriate treatment during interaction. Spencer-Oatey distinguishes between equity rights, which concern fairness, consideration, and freedom from exploitation, and association rights, which reflect individuals' expectations regarding social involvement, inclusion, companionship, and group belonging (see also Spencer-Oatey & Wang, 2026). Communication is perceived as successful when these rights are respected and maintained. However, violations of sociality rights often result in interpersonal tension, relational conflict, or communicative breakdown (see e.g., Caspi & Raz, 2025).

In our study, RMT provides the principal framework for explaining how emojis contribute to the maintenance of interpersonal relationships in ADD. Whereas PT primarily explains the mitigation of face threatening acts, RMT captures broader forms of relational work involving respect, solidarity, empathy, and encouragement. The 2 theories complement each other in tackling IPFs; while PT explains how communicative actions protect face, RMT verifies how those actions collectively establish, maintain, and strengthen social relationships over time (cf. Haugh, 2013). From a RMT perspective, once more, our example of the utterance "*Sorry, could you move a little?* "to allow an interlocuter to pass" could invoke an interpersonal relation, maintaining rapport and avoiding face threat, for instance.

### 2.2.4 Research gap

The above review reveals some major gaps in the existing literature including: i) most computational studies conceptualize emojis primarily as indicators of sentiment or emotion, overlooking their broader interpersonal and relational functions, ii) although digital pragmatic studies have convincingly demonstrated that emojis perform important interactional work (see e.g., Gibson et al., 2018), these theoretical insights have rarely been translated into computational classification frameworks, iii) existing Transformer-based studies overwhelmingly target semantic or affective classification tasks (see e.g., Acheampong et al., 2021) with very limited attention devoted to modeling pragmatically motivated interpersonal functions derived from established linguistic theories such as Brown and Levinson's (1987) PT or Spencer-Oatey's (2000) RMT, and iv) research investigating interpersonal emoji pragmatics in ADD remains virtually absent, despite Facebook representing one of the most widely used digital platforms in the Arab world and MARBERT providing a highly suitable architecture for modeling vernacular Arabic social-media language. Thus, this study aims to bridge these gaps.

## 3. Study design

### 3.1. Data collection and tools

We collected our data from Facebook employing Python 3. The total data collected amounted to 10248 posts. We found 1601 posts without emojis, and 143 posts were duplicates. Both we did not consider in our corpus. Our net data consisted of unique 8504 posts, which were divided into 3 datasets: training dataset containing 4000 posts, validation dataset 2504, and testing dataset 2000. It should be noted here that we did not clean or refine the data. Put differently, we kept hashtags, #es, www.facebook..., https://..., and some other noise ingredients, as these are part of the digital environment. What we excluded are all identifying elements, to adhere to Facebook privacy conditions, and those of the Facebookers. In the case of account names or hashtags/#es, if any contains a name or account, we removed these from them.

### 3.2. Annotation procedure

We adopted a multi-label framework for annotating the unique 8504 posts. Each Facebook post was independently evaluated for the presence or absence of 1-5 IPFs, namely *politeness, respect, solidarity, empathy,* and *encouragement*. Each function was annotated as a binary label (1 = present; 0 = absent), allowing a single emoji or emoji sequence to realize multiple interpersonal functions simultaneously. We recruited 3 annotators, training them for 10 days, 3 hours a day. We gave them the following annotation instructions as protocol:

1. Read the complete Facebook post in its original form to understand the overall communicative context
2. Identify the target emoji(s) and interpret their communicative role within the surrounding textual context rather than in isolation
3. Evaluate the post against all 1-5 interpersonal functions and assign binary labels (1/0 (1 = present; 0 = absent) for each applicable category
4. Base your annotation decisions on contextual meaning, communicative intention, and interpersonal function (please don't look at literal or sentiment-based emoji meanings)
5. Resolve annotation disagreements through discussion and majority voting to produce the final gold-standard dataset.
6. Leave those cases of annotation disagreements that you did not agree on for us, which we will discuss with you and make collective decisions).

The annotation process took 24 days (without training days), and 1 day was spent on resolving annotation disagreements that were left undone (cf. 6).

### 3.3. Data preprocessing and tokenization

All Facebook posts were standardized to enforce uniform UTF-8 text encoding while handling missing fields. Emojis were intentionally preserved in their raw UNICODE textual representations without normalization or translation into descriptive text strings, preserving their natural surface syntax. Each Facebook post was tokenized using the MARBERT tokenizer and represented as input IDs and attention masks, while the 5 IPFs were encoded as independent binary labels. The annotated corpus was partitioned into 3 predefined subsets comprising 4000 posts for training, 2504 posts for validation, and 2000 posts for evaluation/testing. Data partitioning was executed via multi-label stratified sampling (random seed = 42) to preserve the proportional distribution of all 5 pragmatic functions across all splits, minimizing sampling bias. To prepare the input representations for Transformer finetuning, sequence truncation and dynamic padding were applied up to a fixed maximum sequence length ceiling (max = 128).

### 3.4. Model finetuning and optimization

Our study adopts a supervised machine learning pipeline, experimented on Google Colab. MARBERT was finetuned for multi-label classification using the HuggingFace Trainer framework with a PyTorch backend. During training, the model learned to predict the probability of each pragmatic function simultaneously using a binary cross-entropy loss, allowing multiple interpersonal functions to be assigned to the same post. Training was performed for 5 epochs using the AdamW optimizer under mixed-precision (fp16) computation on a Tesla T4 GPU. Model performance was evaluated on the validation set after each epoch, and the checkpoint achieving the highest Macro F1-score was selected as the final model. This model was subsequently evaluated on the independent (unseen) test set to assess its generalization performance.

To elaborate, given that the annotation scheme is inherently multi-label, MARBERT performance was evaluated using complementary metrics that capture different aspects of classification performance. Exact match accuracy was used to measure the proportion of posts whose complete label sets were predicted correctly, while Hamming Loss measured the proportion of incorrectly predicted labels across all 5 interpersonal functions. Micro metrics aggregate true positives, false positives, and false negatives across all labels, providing an overall measure that reflects the model's global classification performance. Given also that the annotation scheme is inherently multi-label, a single Facebook post or emoji may simultaneously express multiple IPFs. Micro Precision, Recall, and F1-score were thus computed to evaluate the model's overall ability to identify these overlapping interpersonal functions across the entire dataset. However, Macro metrics calculate the arithmetic mean of the performance obtained for each interpersonal function, assigning equal importance to every label regardless of its frequency. It follows that macro F1 was adopted as the primary evaluation metric because it provides a more balanced assessment for the imbalanced distribution of IPFs. Weighted metrics additionally account for label frequency by weighting each function according to its support, thereby reflecting overall performance while considering class imbalance. Fig 1 displays part of the Python script code we executed on Colab.

```
File  Edit  Format  Run  Options  Window  Help
!pip install -q transformers datasets accelerate evaluate

from google.colab import files
import pandas as pd

print("[REDACTED]")
uploaded = files.upload()

train_df = pd.read_excel('[REDACTED]')
val_df = pd.read_excel('[REDACTED]')
test_df = pd.read_excel('[REDACTED]')

print("Datasets successfully loaded!")
print(f"Training shape: {train_df.shape}")
print(f"Validation shape: {val_df.shape}")
print(f"Test shape: {test_df.shape}")

import pandas as pd
from datasets import Dataset
from transformers import AutoTokenizer, AutoModelForSequenceClassification, Trainer, TrainingArguments
import torch
import torch.nn as nn
import numpy as np
from sklearn.metrics import accuracy_score, precision_score, recall_score, f1_score, classification_report
```

***Fig 1: Part of Python code***

### 3.5. Operationalizing interpersonal functions

Given the theoretical operationalization sketched in section 2, we conceptualize emojis as multimodal resources that contribute to contextual interpretation, employing SAT, PT and RMT, there should be an operationalizing procedure to enable computational modeling. Thus, in our study emojis are interpreted as multifunctional interpersonal pragmatic resources whose communicative contribution depends upon contextual inference and relational intention. By translating established theories from pragmatics into operational annotation categories, the framework establishes a principled connection between linguistic theory and computational modeling, enabling Transformer-based language models, specifically MARBERT to learn theoretically grounded IPFs directly from authentic ADD. Table 1 summarizes our operationalization framework: (see also Austin, 1962; Brown & Levinson, 1987; Spencer-Oatey, 2000, 2008; Culpeper, 2011; Haugh, 2013; Locher & Watts, 2005).

**Table 1: Operationalization framework**

| Pragmatic function | Speech act | Emoji example |
|---|---|---|
| **Politeness** | Softening directives, buffering face-threats, signaling courtesy | 💙, 😊, 🙏 |
| **Respect** | Expressing formal deference, appreciation, acknowledging standing | 🌹, 👏, 💐 |
| **Solidarity** | Constructing in-group cohesion, affiliation, and social closeness | 🤝, 🤗, 💯 |
| **Empathy** | Displaying emotional resonance, compassion, and support | 😢, 🥺, 💔 |
| **Encouragement** | Providing motivation, validation, and positive reinforcement | 💪, 👍, 🔥 |

### 3.6 Methods of analysis

MARBERT was evaluated using standard metrics for multi-label text classification. Since each Facebook post may simultaneously express more than one interpersonal pragmatic function, the evaluation considered both micro-averaged and macro-averaged measures to provide complementary perspectives on model performance. In addition, overall accuracy, Hamming loss, and weighted F1-score were computed. The metrics were calculated from the numbers of TP (=true positives), TN (=true negatives), FP (=false positives), and FN (=false negatives). As for macro and accuracy metrics, they were calculated as follows:

$$Accuracy = \frac{TP + TN}{TP + TN + FP + FN}$$

$$Hamming\ Loss = \frac{1}{N \cdot |Y|} \sum_{i=1}^{N} \sum_{c \in Y} I\left(y_{i,c} \neq \widehat{y_{i,c}}\right)$$

$$Macro\ Precision = \frac{1}{|Y|} \sum_{c \in Y} \frac{TP_c}{TP_c + FP_c}$$

$$Macro\ Recall = \frac{1}{|Y|} \sum_{c \in Y} \frac{TP_c}{TP_c + FN_c}$$

$$Macro\ F1 - Score = \frac{1}{|Y|} \sum_{c \in Y} \frac{2 \cdot Precision_c \cdot Recall_c}{Precision_c + Recall_c}$$

$$Weighted\ F1 - Score = \sum_{c \in Y} \frac{n_c}{N} F1_c$$

However, micro metrics were computed as follows:

$$Micro\ Precision = \frac{\sum_{c \in Y} TP_c}{\sum_{c \in Y}(TP_c + FP_c)}$$

$$Micro\ Recall = \frac{\sum_{c \in Y} TP_c}{\sum_{c \in Y} TP_c + \sum_{c \in Y} FN_c}$$

$$Micro\ F1\text{-}Score = \frac{2 \cdot Micro\ Precision \cdot Micro\ Recall}{Micro\ Precision + Micro\ Recall}$$

We amalgamated this statistical approach with qualitative approaches utilizing PT, SAT, and RMT. These provide the theoretical framework in which we interpret the computational and statistical findings.

**4. Results**

In this section, we present the study results focusing mainly on accuracy, micro and macro metrics as Table 2 demonstrates.

**Table 2: MARBERT training-validation performance**

| Epoch | Train.Loss | Valid. Loss | Accuracy | Ma Precision | Ma Recall | Ma F1 | Mi Precision | Mi Recall | Mi F1 |
|---|---|---|---|---|---|---|---|---|---|
| 1 | 0.2949 | 0.2237 | 0.8838 | 0.1253 | 0.0415 | 0.0623 | 0.3171 | 0.0760 | 0.1226 |
| 2 | 0.1853 | 0.1900 | 0.9081 | 0.6759 | 0.1515 | 0.2275 | 0.7672 | 0.2602 | 0.3886 |
| 3 | 0.1416 | 0.1595 | 0.8786 | 0.4256 | 0.5045 | 0.4503 | 0.4779 | 0.5994 | 0.5318 |
| 4 | 0.0802 | 0.1375 | 0.9301 | 0.6837 | 0.4897 | 0.5555 | 0.7333 | 0.5789 | 0.6471 |
| 5 | 0.0478 | 0.1337 | 0.9329 | 0.7722 | 0.5690 | 0.6531 | 0.7575 | 0.5936 | 0.6656 |
| Testing | -- | -- | 0.9290 | 0.7300 | 0.4600 | 0.5617 | 0.7540 | 0.5184 | 0.6144 |

The training trajectory indicates stable optimization, with both training and validation losses decreasing consistently across epochs. The increase in macro F1 from 0.06 in epoch 1to 0.65 in epoch 5 demonstrates that MARBERT progressively improved its ability to recognize multiple interpersonal functions, including less frequent categories. Epoch 5 achieved the strongest validation performance, suggesting successful adaptation of the pretrained representations to the emoji pragmatic classification task. The training trajectory indicates stable optimization, with both training and validation losses decreasing consistently across epochs. The increase in macro F1 from 0.06 in the first epoch to 0.65 in the final epoch demonstrates that MARBERT progressively improved its ability to recognize multiple interpersonal functions including less frequent categories. Epoch 5 achieved the strongest validation performance, suggesting successful adaptation of the pretrained representations to the emoji pragmatic classification task. The model's performance on the latter results are presented in Table 3.

**Table 3: Function-level analysis & Avgs**

| Pragmatic function | Precision | Recall | F1-score | Support |
|---|---|---|---|---|
| **Politeness** | 0.78 | 0.57 | 0.66 | 126 |
| **Respect** | 0.82 | 0.52 | 0.64 | 54 |
| **Solidarity** | 0.75 | 0.30 | 0.43 | 10 |
| **Empathy** | 0.71 | 0.47 | 0.56 | 51 |

| Encouragement | 0.61 | 0.45 | 0.52 | 31 |
|---|---|---|---|---|
| **Macro Avg** | 0.73 | 0.46 | 0.56 | 272 |
| **Micro Avg** | 0.75 | 0.52 | 0.61 | 272 |
| **Weighted Avg** | 0.76 | 0.52 | 0.61 | 272 |
| **Hamming Loss** | 0.0169 | | | |

Table 3 presents the function-level analysis revealing meaningful variation in MARBERT's ability to identify different IPFs. Politeness achieved the highest performance, with an F1-score of 0.66, followed closely by Respect with an F1-score of 0.64. Empathy and Encouragement achieved moderate performance with F1-scores of 0.56 and 0.52, respectively, while Solidarity recorded the lowest performance at F1 0.43 (with a low recall of 0.30). Regarding the aggregate metrics, the Macro Avg F1 0.56 calculates the unweighted mean across all classes, whereas the Weighted Avg F1 0.61 accounts for class imbalance by scaling scores by support. The Micro Avg yields an F1-score of 0.61, aggregating the total contributions across all categories.

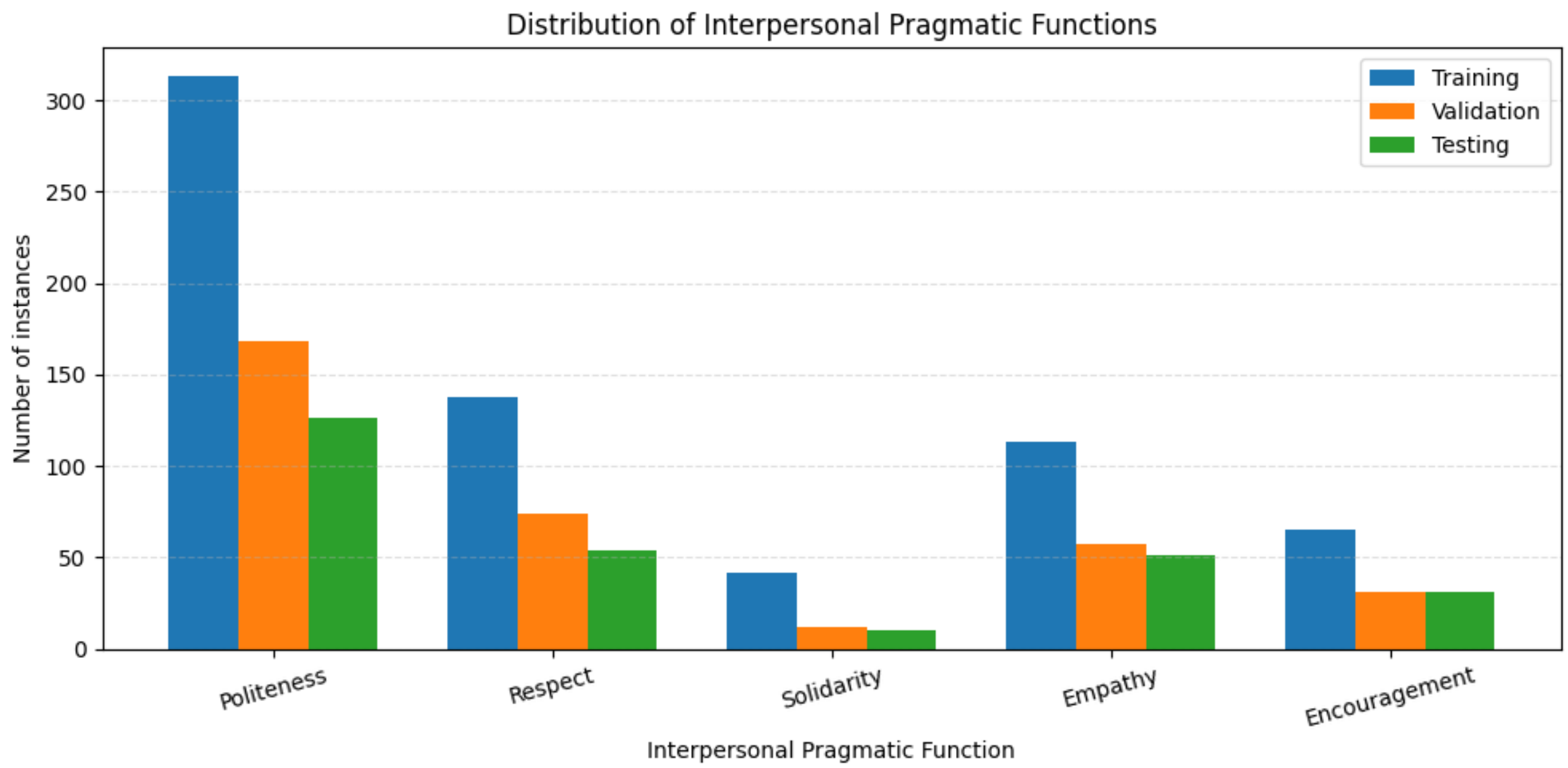

***Fig 2: Distribution of interpersonal pragmatic functions***

Fig 2 illustrates the dataset split distribution, divided into training, validation, and testing sets across 5 interpersonal pragmatic functions. It is clear that *politeness* is the most heavily represented function, followed by *respect* and *empathy* as moderately frequent categories, while *encouragement* and *solidarity* appear less frequently, with solidarity representing the smallest category overall. Across all categories, the splits maintain a consistent proportional hierarchy of training, validation, and testing sets, reflecting a standard machine learning partition strategy for text classification modeling. This suggests that the difficulty encountered by MAREBRT in modeling the 5 pragmatic functions significantly differs. This is also consolidated by confusion matrices as Fig 3 showcases.

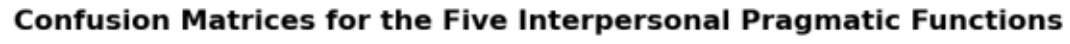

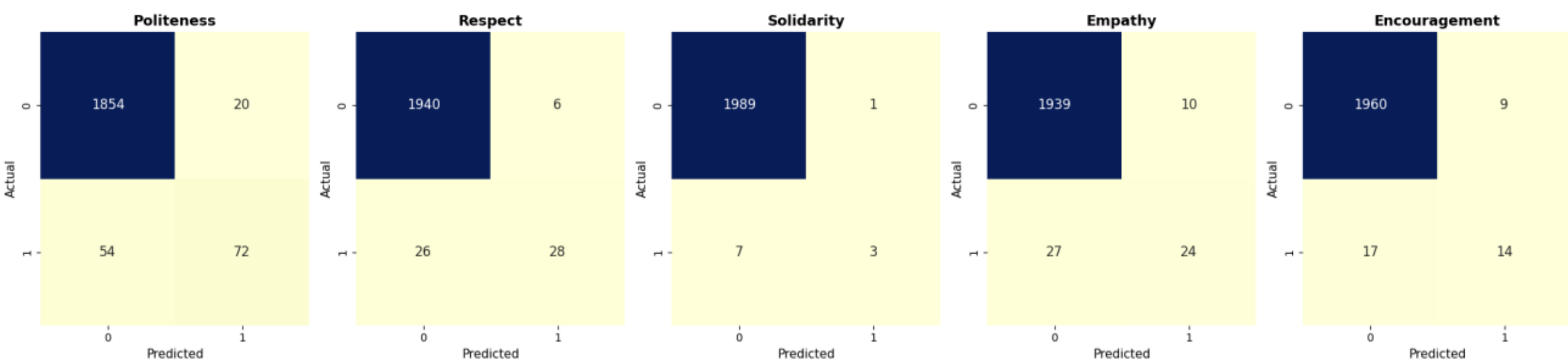

***Fig 3: Confusion Matrices***

Fig 3 presents the confusion matrices for the 5 pragmatic functions and provides insight into the types of classification errors produced by MARBERT. The matrices indicate that the model achieved relatively strong discrimination for *politeness* and *respect*, which correspond to the highest-performing categories in the function-level evaluation. These functions appear to contain more explicit contextual cues, allowing the model to establish stronger associations between linguistic context and interpersonal intent. However, greater confusion is observed among categories involving relational meanings, particularly *solidarity*, *empathy*, and *encouragement*. This overlap reflects the pragmatic flexibility of emojis in ADD. For example, an emoji expressing solidarity or encouragement may simultaneously signal empathy towards another user, solidarity with a group, or encouragement for future action. Thus, some classification uncertainty is expected because these categories represent related interpersonal functions rather than completely independent semantic classes.

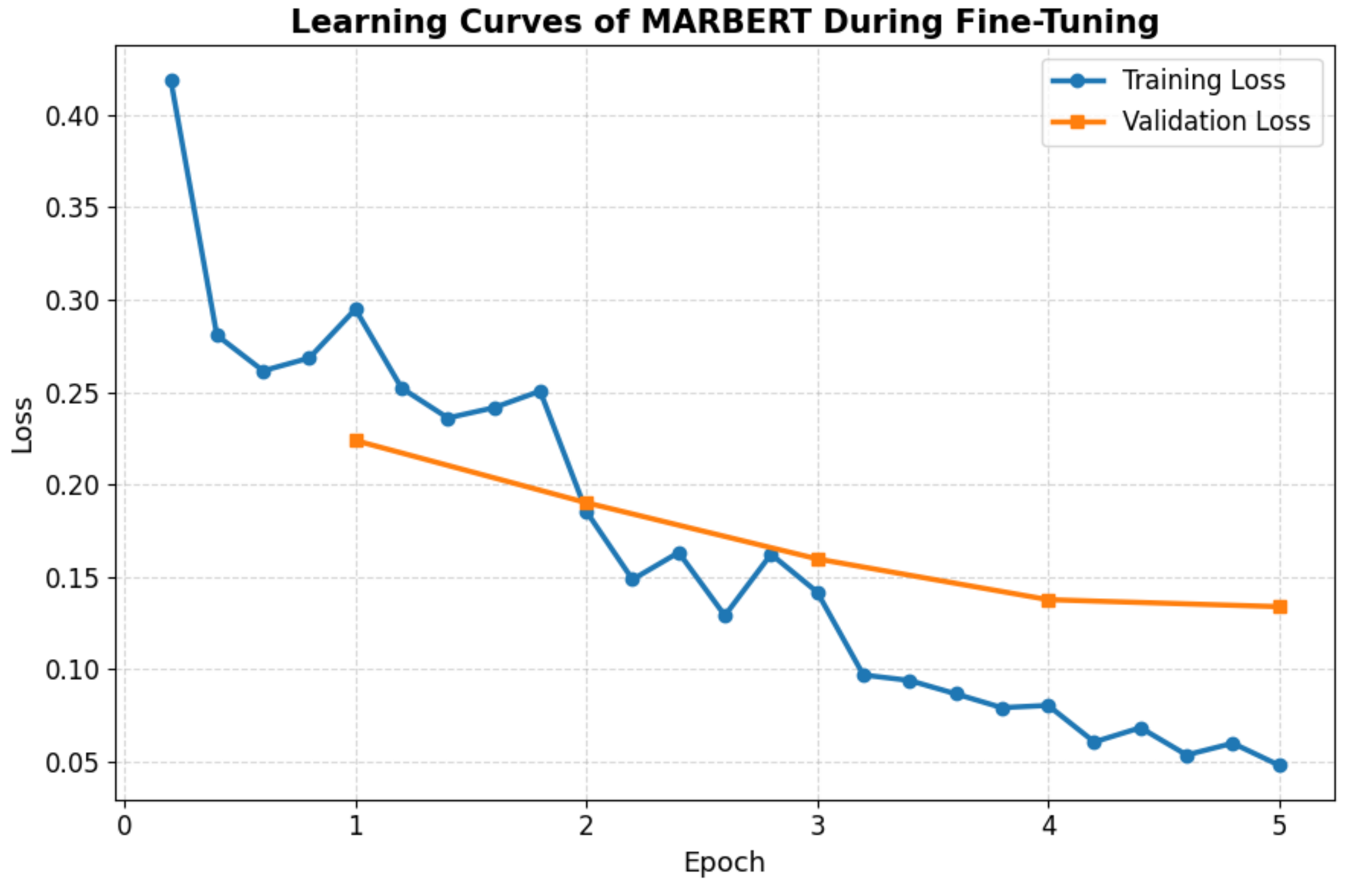

***Fig 4: Learning curves***

As Fig 4 shows, the learning curves illustrating the MARBERT training and validation losses during finetuning demonstrate a steady optimization process across the 5 epochs. Training loss was logged at multiple optimization steps within each epoch, whereas validation loss was computed once at the end of each epoch (cf. Table 2). The training loss degrades from ≈ 0.295 by (the end of) epoch 1 and exhibits a downwards trend with minor fluctuations, eventually descending to ≈ 0.05 by epoch 5. Similarly, the validation loss exhibits a consistent decline from roughly ≈ 0.22 at (the end of) epoch 1 and downwards to ≈ 0.13 by epoch 5. The parallel downwards trajectory of both curves indicates effective learning and generalization without severe, unchecked overfitting, as the validation loss stably decreases alongside the training loss.

## 5. Discussion

Given our findings above, the evaluation on the independent test dataset confirms MARBERT's ability to generalize effectively to unseen Facebook posts. MARBERT achieved an overall accuracy of 93% with a Hamming loss of 0.02, indicating that only a small proportion of label assignments were incorrect despite the complexity of the multi-label classification task. These results demonstrate that transformer-based contextual representations successfully capture the linguistic and pragmatic regularities underlying emoji use in AFD. The low Hamming loss further suggests that MARBERT consistently predicts multiple interpersonal functions simultaneously, an important requirement given that emoji meanings frequently overlap rather than forming mutually exclusive categories. Given that the 5 IPFs are unevenly distributed across the corpus, overall accuracy alone cannot adequately reflect model performance. Macro-averaged measures therefore provide a more balanced evaluation by assigning equal importance to each pragmatic category regardless of frequency.

MARBERT behavior on the unseen data achieves a macro precision of 0.73, macro recall of 0.46, and a macro F1-score of 0.56, demonstrating quite robust performance across both frequent and infrequent pragmatic categories despite substantial class imbalance (for training and validation performance, see Table 2). Although the comparatively lower macro recall reflects the difficulty of identifying less represented functions, particularly Solidarity, the macro F1-score indicates that MARBERT successfully captures meaningful pragmatic distinctions beyond the dominant categories. The difference between the weighted and macro scores reflects the inherent challenge of detecting less represented and more context-dependent interpersonal functions, while simultaneously demonstrating that the model learns conventionalized pragmatic patterns more effectively when sufficient training examples are available.

The micro-level performance on testing/evaluation provides additional evidence of MARBERT's ability to capture recurrent interpersonal patterns across the corpus. The model achieves a micro precision of 0.75, micro recall of 0.52, and micro F1-score of 0.61, indicating that it successfully learns systematic contextual associations between linguistic expressions and emoji functions when predictions are aggregated across all instances. The higher micro-level performance reflects MARBERT's effectiveness in modeling recurrent pragmatic cues that characterize naturally occurring Facebook interactions (cf. van der Vlist et al., 2022). Comparatively, the micro-average metrics (precision = 0.75, recall = 0.52, F1 = 0.61) demonstrate improved overall performance when predictions are aggregated across all instances, reflecting the model's effectiveness in capturing recurrent interpersonal patterns in the corpus. A similar trend is observed in the weighted-average scores (precision = 0.76, recall = 0.52, F1 = 0.61) (cf. Table 3), which give greater influence to the more frequent pragmatic functions. All in all, the macro, micro, and average level metrics suggest that MARBERT performs particularly well for interpersonal functions characterized by conventionalized linguistic realizations while remaining challenged by functions that rely primarily on implicit contextual knowledge, shared identity, and discourse-dependent interpretation.

The performance hierarchy observed across the functional categories reveals a clear structural division between explicit interpersonal routines and implicit relational constructs. Politeness and Respect achieved superior performance, recording F1-scores of 0.66 and 0.64, respectively. To elaborate, Politeness achieves precession 0.78, recall 0.57, and F1 score 0.66, indicating that MARBERT robustly identifies a high level of interpersonal function correctness supported by 126 instances. However, the particularly high precision for Respect 0.82 indicates that MARBERT effectively identifies contexts in which emojis function as markers of appreciation, recognition, or deference. These high scores stem from their frequent realization through conventionalized communicative practices, where emojis accompanying expressions of gratitude, acknowledgment, greeting, or positive evaluation consistently co-occur with recognizable lexical markers. The differences reflect the varying degrees of explicitness with which interpersonal meanings are encoded in naturally occurring Facebook interactions (cf. Figs 2 & 3). Similar observations were reported by Caspi and Raz (2025), who demonstrated that emojis function not merely as affective markers but as interactional resources that soften utterances, reinforce positive interpersonal attitudes, and strengthen rapport between interlocutors. Recent work in emoji pragmatics emphasizes that emojis derive much of their communicative value from their ability to regulate interpersonal relationships rather than simply conveying emotional states (cf. Ge & Herring, 2018; Gibson et al., 2018; Yus, 2025).

These findings are readily interpretable within Brown and Levinson's (1987) PT. To repeat a point stated earlier, PT proposes that speakers employ linguistic and paralinguistic strategies to preserve both positive and negative face during social interaction. Positive face concerns an individual's desire to be appreciated, approved of, and socially accepted, whereas negative face relates to the desire for autonomy and freedom from imposition. Throughout the Facebook corpus examined in this study, emojis frequently accompany expressions of gratitude, praise, congratulations, acknowledgment, encouragement, and respect, functioning primarily as positive politeness strategies that reduce interpersonal distance, emphasize shared goodwill, and reinforce harmonious social relationships. Because these communicative acts exhibit relatively stable lexical realizations and predictable discourse environments, they generate consistent contextual representations that MARBERT successfully learns. The model's strong performance therefore reflects not only its computational capacity but also the linguistic regularity with which positive facework is realized in naturally occurring AFD.

While Brown and Levinson's (1987) PT explains the prominence of face-saving strategies, Spencer-Oatey's RMT (2000, 2008) offers a broader account of the observed performance differences by extending interpersonal pragmatics beyond face alone. Rapport Management distinguishes between quality face, identity face, sociality rights, and interactional goals, thereby capturing the multidimensional nature of interpersonal communication. The high performance obtained for Politeness and Respect largely reflects quality face, whereby speakers recognize one another's competence, worth, achievements, or social value through highly conventionalized linguistic expressions accompanied by supportive emojis. These communicative acts are recurrent, structurally predictable, and thus computationally learnable. They also reinforce sociality rights by signaling consideration, reciprocity, and appropriate interpersonal conduct, thereby contributing to the maintenance of harmonious online interaction. However, identity face presents considerably greater challenges for computational modeling because it depends less on explicit linguistic realization than on shared group membership, collective experiences, ideological affiliation, interpersonal history, and culturally situated background knowledge. These dimensions are rarely expressed directly within isolated Facebook posts and instead emerge through accumulated social interaction and contextual inference. Thus, the finetu between the robust performance observed for Politeness and Respect and the weaker performance for Solidarity mirrors the theoretical distinction proposed by RMT between relatively explicit forms of facework and more implicit forms of rapport construction. In this respect, MARBERT appears to reproduce the same hierarchy of interpersonal explicitness predicted by pragmatic theory: communicative functions grounded in overt linguistic realization are learned more successfully than those requiring extensive social and contextual inference.

This distinction becomes particularly evident in the case of *solidarity*, which achieved the lowest performance (F1-score of 0.43) and emerged as the most challenging interpersonal pragmatic function. Unlike *politeness* or *respect*, *solidarity* is rarely encoded through explicit lexical cues alone (cf. Table 2). It is constructed, instead, through shared identity, common experiences, collective memory, interpersonal history, ideological positioning, and mutual social affiliation. Additionally, the inherent multifunctionality and ambiguity of emoji use, where identical symbols signal varied pragmatic intentions depending on discourse positioning, contributed to classification overlap. It then follows that identical emojis may communicate solidarity in one interaction while expressing empathy, encouragement, or even simple emotional alignment in another, depending entirely on the surrounding discourse and shared contextual assumptions. This observation agrees with Hand et al.'s (2022) study which demonstrated that emoji interpretation depends heavily on the interaction between textual content and emoji type, with identical symbols yielding different pragmatic meanings across discourse contexts. Similarly, Tieu et al. (2025) argue that emojis trigger flexible pragmatic inferences whose interpretation varies according to discourse position and communicative intention, while Weissman (2022) maintains that emoji meaning constitutes a fundamentally pragmatic rather than semantic phenomenon. In this sense, the classification overlap observed among Solidarity, Empathy, and Encouragement should not be interpreted merely as model error but rather as computational evidence of the inherently multifunctional nature of interpersonal meaning.

Beyond the computational and theoretical explanations, the distribution of IPFs also invites cautious reflection on the broader sociocultural environment in which current Arabic digital discourse is situated. Although the relatively small number of Solidarity instances primarily reflects the characteristics of the annotated corpus and should not be interpreted as a direct representation of Arab society, the findings resonate with wider patterns documented in recent studies of Arabic online communication identities (see e.g., Shormani & Alenezi, 2026). Current ADD increasingly unfolds against a backdrop of prolonged regional conflicts, political instability, forced displacement, economic uncertainty, ideological polarization, and fragmented national and transnational identities (Shormani & Alenezi, 2026). These conditions have reshaped patterns of interpersonal communication, often encouraging users to express individual sympathy, respect, or encouragement towards particular interlocutors while making broader expressions of collective solidarity comparatively less frequent and more context-dependent. Shormani and Alenezi provide a corpus pragmatics study examining Arabic nominalization in digital discourse. They argue that grammatical choices frequently perform pragmatic actions that reflect evolving social identities, ideological positioning, and collective attitudes rather than serving purely structural functions. Their findings suggest that contemporary Arabic digital discourse increasingly indexes fragmentation, negotiation of identity, and competing social affiliations, while simultaneously revealing a persistent orientation towards shared cultural memory and collective belonging.

Within this broader sociopragmatic landscape, solidarity emerges not as a routinely expressed interpersonal strategy but as a more selective and contextually negotiated communicative resource, activated primarily when speakers explicitly invoke shared identities, common causes, or collective experiences. As supported by recent scholarly discourse, contemporary Arab societies frequently experience a state of divide characterized by a lack of systemic harmony and heightened individualism. This fragmented social reality is empirically mirrored in the sparse representation of supportive interactions (10 support instances in evaluation dataset), rendering the computational modeling of *solidarity* inherently elusive. At the same time, Arab digital communication continues to exhibit a profound sense of collective memory and nostalgia, frequently expressed through references to shared history, cultural heritage, religious rituals, linguistic identity, and memories of periods perceived as socially or politically more brilliant. Nostalgia functions not merely as remembrance but as a pragmatic resource through which speakers negotiate present uncertainties by invoking an idealized collective past. Expressions of longing for unity, stability, dignity, and common identity frequently coexist with discussions of current crises, producing discourse that simultaneously reflects fragmentation and aspirations for renewed social cohesion (cf. Abu-Nimer et al., 2007). In this sense, emojis participating in supportive interactions often acquire meanings that extend beyond interpersonal intent, indexing broader cultural identities and shared emotional experiences that cannot always be recovered from isolated textual contexts alone. Such discourse further illustrates why identity-based interpersonal functions remain substantially more difficult for contextual language models to learn than conventionalized expressions of politeness or respect.

To recapitulate, the computational findings extend beyond model evaluation to provide empirical support for contemporary theories of interpersonal pragmatics. MARBERT does more than accurately classify emoji functions; it reproduces a meaningful hierarchy of interpersonal explicitness predicted by both PT and RMT. Conventionalized face-management strategies such as politeness and respect, are learned more reliably because they are realized through stable linguistic and interactional patterns. However, relational functions such as solidarity depend on implicit identity construction, collective memory, shared sociocultural knowledge, and contextual inference, making them inherently more difficult for both humans and machines to interpret. It follows then that these findings demonstrate that the computational learnability of emoji pragmatics closely mirrors the degree of interpersonal explicitness with which rapport, facework, identity, and social relationships are constructed in ADD, thereby strengthening the intersection between computational linguistics, interpersonal pragmatics, and digital communication.

## 6. Conclusion and limitations

This study investigated the feasibility of modeling the IPFs of emojis using the state-of-the-art MARBERT. Focusing on AFD, the study proposed a multi-label classification framework that distinguished 5 interpersonal functions, namely *politeness, respect, solidarity, empathy,* and *encouragement,* and evaluated the performance of MARBERT in learning these context-dependent pragmatic meanings. The results demonstrate that Transformer-based language models can successfully capture important aspects of emoji pragmatics beyond traditional sentiment or emotion classification. Specifically, MARBERT achieved strong overall performance in recognizing interpersonal functions associated with conventionalized linguistic patterns and quality face, while also revealing the greater computational complexity of functions like solidarity and identity face, whose interpretation relies heavily on implicit social context, collective memory, and relational inference.

Beyond its computational contribution, the study advances the growing intersection between interpersonal pragmatics and NLP. Treating emojis as pragmatic resources rather than merely affective symbols, the study demonstrates that contextual language models are capable of approximating higher-order communicative intentions embedded in naturally occurring online interaction. The findings provide empirical support for theoretical perspectives such as Brown and Levinson's (1987) PT and Spencer-Oatey's (2000) RMT by showing that many interpersonal functions expressed through emojis exhibit systematic contextual regularities that can be learned computationally. At the same time, the comparatively lower performance for solidarity highlights the inherently contextual, dynamic, and socially constructed nature of interpersonal meanings, reflecting wider sociopragmatic realities such as current fragmentation and social divide, and emphasizing that not all pragmatic phenomena are equally accessible through textual information alone. Methodologically, this study contributes one of the first Transformer-based multi-label classification frameworks, specifically designed to model IPFs of emojis in Arabic social media discourse, distinguished from previous computational research that has treated emojis primarily as indicators of sentiment or emotion (see e.g., Kralj Novak et al., 2015; Abdul-Mageed & Elmadany, 2021; Abdul-Mageed et al., 2022; Ahamad & Nishra, 2025; Zhao et al., 2018).

Despite these contributions, several limitations should be acknowledged here: i) the corpus was restricted to Arabic Facebook posts, which represent only one social media environment. Emoji use may vary considerably across platforms such as X, Instagram, and/or Telegram because each platform supports different interactional norms, audience expectations, and communicative practices, ii) although the dataset represents one of the few manually annotated corpora targeting interpersonal emoji functions, the distribution of labels remained naturally imbalanced. Some pragmatic functions, particularly Solidarity, occurred much less frequently than Politeness or Respect, iii) this study relies exclusively on textual context and emojis. However, interpersonal meaning in social media communication is often influenced by additional contextual factors, including conversational history, speaker relationships, shared background knowledge, multimodal content, images, videos, and platform-specific interactional features. Future studies could focus on incorporating these broader sources of contextual information, and iv) the study evaluated only MARBERT. Although MARBERT proved highly effective for the study task, future research may involve evaluating other LLMs and multilingual Transformer architectures to provide further insights into how pretrained language models encode pragmatic knowledge and interpersonal relations.

**Availability of data and materials**
The training, validation, and evaluation datasets generated or analyzed during the current study will be available on GitHub at [https://github.com/..../ marbert_emoji_modeling...]

**Funding**
This research project did not receive internal or external funding.

**Conflicts of Interest**
Authors declare no conflicts of interest.